\documentclass[runningheads]{llncs}
\usepackage[T1]{fontenc}
\usepackage{graphicx}
\usepackage{amsmath,amssymb,amsfonts}
\usepackage{algorithmic}
\usepackage{algorithm}
\usepackage{subfigure}
\usepackage{color}
\usepackage{textcomp}

\begin{document}
\title{GAMDTP: Dynamic Trajectory Prediction with Graph Attention Mamba Network}
%
%
\author{Yunxiang Liu\inst{1} \and
Hongkuo Niu\inst{1} \and
Jianlin Zhu\inst{2}
}
\authorrunning{L. YunXiang, N. HongKuo, Z. JianLin}
%
\institute{Shanghai Institute of Technology, Shanghai 10259, China\\
\email{236142132@mail.sit.edu.cn}\\
}

\maketitle              
\begin{abstract}
    Accurate motion prediction of traffic agents is crucial for the safety and stability of autonomous driving systems. In this paper, we introduce GAMDTP, a novel graph attention-based network tailored for dynamic trajectory prediction. Specifically, we fuse the result of self attention and mamba-ssm through a gate mechanism, leveraging the strengths of both to extract features more efficiently and accurately, in each graph convolution layer. GAMDTP encodes the high-definition map(HD map) data and the agents' historical trajectory coordinates and decodes the network's output to generate the final prediction results. Additionally, recent approaches predominantly focus on dynamically fusing historical forecast results and rely on two-stage frameworks including proposal and refinement. To further enhance the performance of the two-stage frameworks we also design a scoring mechanism to evaluate the prediction quality during the proposal and refinement processes. Experiments on the Argoverse dataset demonstrates that GAMDTP achieves state-of-the-art performance, achieving superior accuracy in dynamic trajectory prediction.

\end{abstract}
\section{Introduction}
Accurate motion forecasting of surrounding traffic agents, including vehicles, pedestrians, and other road participants, is critical to guarantee the safety and stability of autonomous driving systems. Predicting the trajectories of traffic agents with high precision allows autonomous systems to anticipate future states, make informed decisions in real-time and avoid risks while driving.\par
Researches in the early stage mainly used rasterized segmantic images to represent map information\cite{lee2017desire,phan2020covernet}. However, due to the loss of information while rasterization, \cite{gao2020vectornet} and \cite{liang2020learning} both design a vector-based method that agents and roads are modeied as a collection of vectors. \cite{wang2020multi,azadani2023stag,zhang2022trajectory,chen2022intention} are based on this and leverage GNNs\cite{velickovic2017graph} and LSTM\cite{hochreiter1997long} to fuse spatio-temporal information for accurate and socially plausible vehicle trajectory prediction. However, LSTM-based methods are bottlenecked by the parallelization, memory efficiency, long term dependencies and training speed. Recent advances in this domain such as HiVT\cite{zhou2022hivt}, by considering the deep relationship between agents and scenario, agents and agents, as well as the selection of the direction of the coordinate system and other factors, the network achieves a fairly good effect. QCNet\cite{zhou2023query} further investigates the impact of reusing historical calculations on the final prediction results. They presents an efficient, multi-modal trajectory prediction framework using a novel tow-stage, consists of proposal and refinement, with query-centric paradigm. By reusing scene encodings and combining anchor-based refining strategies, it achieves both fast inference and high prediction accuracy, making it well-suited for real-time autonomous driving scenarios. Morever, HPNet\cite{tang2024hpnet} integrates historical predictions with real-time context through its Historical Prediction Attention module, which dynamically models the relationship between successive predictions, resulting in more accurate and stable trajectory forecasts. In addition, many previous works\cite{chai2019multipath,gu2021densetnt,liu2021multimodal,varadarajan2022multipath++,zhou2022hivt,zhou2023query,tang2024hpnet} use multi-modal future trajectories as output rather than a single trajectory, given the uncertainty of future, and we also follow this way in this paper.\par

\begin{figure*}[t]
    \centerline{\includegraphics[width=\textwidth,height=0.3\textwidth]{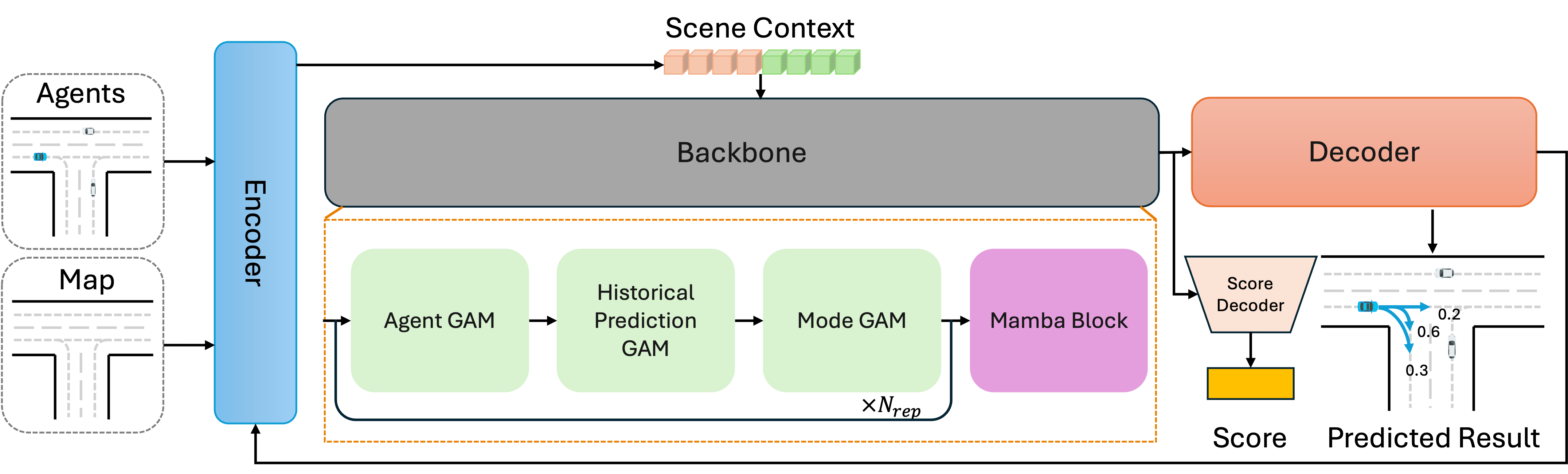}}
    \caption{Overview of GAMDTP. The encoder processes raw input features such as HD map and agent trajectory information. Our proposed Graph Attention Mamba module is applied in the components Agent GAM, Historical Prediction GAM and Mode GAM, which extracts spatio-temporal features. Decoder generates the final predicted trajectories and probability and the score decoder further evaluates and prioritizes trajectory candidates for refinement through generate a score for each result, ensuring accurate and reliable predictions.}
    \label{overview}
\end{figure*}

While most of those approaches are Graph Attention Networks (GAT)\cite{velickovic2017graph} based, which brings GNNs and Transformers together, and Transformers\cite{vaswani2017attention} can capture long range dependencies among nodes in a graph, they suffer from the limitation of quadratic computational complexity due to the self-attention mechanisms, making them less efficient for large-scale, real-time trajectory forecasting tasks. Recently, a brand new state space model (SSM)\cite{gu2021efficiently}, Mamba\cite{gu2023mamba,dao2024transformers}, demonstrates potential in sequence modeling and long-term dependencies capturing with linear computational complexity and improved GPU efficiency across tasks in natural language processing\cite{lieber2024jamba,team2024jamba,he2024densemamba} and computer vision\cite{zhu2024vision,li2025videomamba,zhang2025motion}. Despite its potential, Mamba-SSM remains underexplored in the context of graph-based trajectory prediction frameworks.\par
To address these limitations, we propose GAMDTP, a novel module that fuses Graph Attention Networks (GAT)\cite{velickovic2017graph} with the selective capabilities of Mamba-SSM\cite{gu2023mamba}. Inspired by \cite{ding2024combining} in computational pathology, GAMDTP leverages the unique strengths of both GAT\cite{velickovic2017graph} and Mamba-SSM\cite{dao2024transformers,gu2023mamba} through a gate mechanism, combining the self-attention mechanism’s adaptability to complex inter-agent interactions with Mamba’s efficient handling of long-range dependencies through structured state spaces. This fusion allows GAMDTP to deliver accuracy feature extraction efficiency, scalable computational performance, the ability to adapt to diverse and dynamic driving environments and making it particularly suited for real-time trajectory prediction.\par
Additionally, recognizing the limitations of existing two-stage trajectory prediction frameworks, where the proposal and refinement stages often lack effective cooperation, we introduce a Quality Scoring Mechanism following SmartRefine\cite{zhou2024smartrefine}. This mechanism evaluates the prediction quality at both stages, prioritizing high-quality trajectory proposals and improving the refinement process, ultimately leading to more accurate and reliable trajectory forecasts.\par
Our approach is evaluated on the Argoverse\cite{chang2019argoverse} and INTERACTION\cite{zhan2019interaction} datasets, both are standard benchmarks for autonomous driving scenarios, where GAMDTP demonstrates state-of-the-art performance. This enhancement in prediction capability not only strengthens the robustness of trajectory predictions but also contributes to the overall safety and stability of autonomous driving systems.\par
In summary, our work has the following contributions:
\renewcommand{\labelitemi}{\textbullet}
\begin{itemize}
\item We fuse GAT and Mamba as a novel graph neural network called GAM which combines Mamba module through a gate mechanism to dynamically balance local and global feature extraction.
\item GAMDTP merges a score mechanism to evaluate the prediction results of proposal and refinement to improve the performence of the refine process.
\item Experiments on the Argoverse\cite{chang2019argoverse} and INTERACTION\cite{zhan2019interaction} datasets demonstrate that GAMDTP achieves the state-of-the-art performance.
\end{itemize}

\section{Related work}

\subsection{GNNs and Temporal Models for Trajectory Prediction}
The development of accurate and efficient trajectory prediction models is critical for autonomous driving, as they allow for anticipating the future states of traffic agents ensuring safety and operational stability for real-time decisions. To model the social spatial and temporal interactions between agents and agents, agents and lanes, \cite{liang2020learning,wang2020multi} apply message-passing GNNs and encode agents and lanes as nodes, speed, direction and other dynamic information as edges. GNNs work by iteratively gathering information from neighboring nodes to update the current node’s representation, with different GNN types employing distinct aggregation and update functions. This process enables GNNs to learn representations that encapsulate the graph data’s topological structure. To model history trajectory and other sequence data, early approaches relied heavily on Recurrent Neural Networks(RNNs)\cite{schmidt2019recurrent} and Long Short-Term Memory networks(LSTMs)\cite{hochreiter1997long} to model temporal dependencies in sequential data\cite{zyner2019naturalistic,lee2017desire}. LSTMs have been widely used in autonomous driving applications for their ability to maintain sequential information over time and handle agent-specific histories\cite{xing2019personalized,alahi2016social,deo2018convolutional,chen2022intention}. Compared to LSTMs, Transformers show more powerful in parallelization and long-term dependency capture, which impacts both training and memory efficiency. Therefore, attention mechanism\cite{vaswani2017attention} has become the dominant method adopted by recent\cite{hou2022structural,li2023interaction}. \cite{azadani2023stag,gu2021densetnt,zhou2022hivt,ngiam2021scene,wang2024mm} fuse GNNs and Transformers and model different scenarios toward different cases.\par
Recently, a novel state space model (SSM)\cite{gu2023mamba}, Mamba, has shown promise in sequence modeling and capturing long-term dependencies\cite{gu2021efficiently}. Mamba introduces a selective mechanism into the SSM, enabling it to identify critical information similarly to an attention mechanism. Studies have highlighted Mamba’s potential across domains like natural language processing\cite{lieber2024jamba,team2024jamba,he2024densemamba} and computer vision. However, Mamba’s potential in combination with GATs remains underexplored. In this paper, we fuse Mamba and attention mechanism in graph neural network with a gate mechanism for encoding HD map data and historical trajectory information.\par

\begin{figure}[tbp]
\centerline{\includegraphics[width=0.3\textwidth,height=0.35\textwidth]{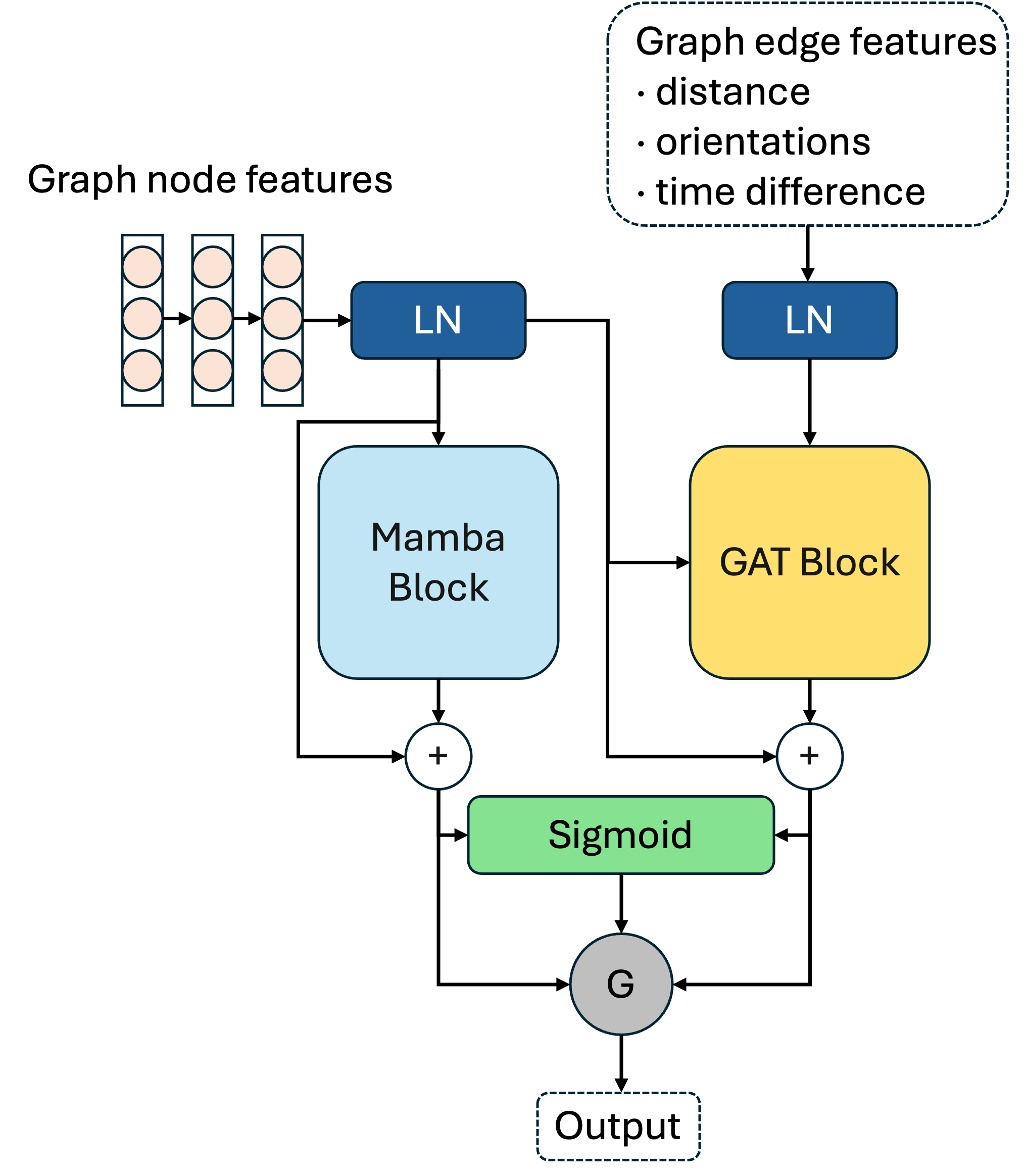}}
\caption{Our proposed Graph Attention Mamba module, which integrates Mamaba block and graph attention block. The input features include node features and edge features, which first normalized through a Layernorm(LN) layer before processed by Mamba and GAT blocks. The output from these blocks are fused using a gate mechanism, where the sigmoid function dynamically generates a gate signal G to balance their contributions.}
\label{GAMpng}
\end{figure}

\subsection{Two-Stage Motion Forecasting}
Inspired by the refinement networks\cite{carion2020end,ren2016faster} in computer vision, refinement strategies have recently been applied in motion forecasting. This framework typically involves a proposal stage, where multiple candidate trajectories are generated, followed by a refinement stage, where these proposals are optimized based on the context. QCNet\cite{zhou2023query} employs a two-stage approach to improve efficiency and accuracy. Specifically, they leverages a query-centric paradigm to forecast the trajectory in the proposal stage and predict the offset in the refinement stage. HPNet\cite{tang2024hpnet} introduces a historical prediction attention module to encode the dynamic relation between successive predictions in the proposal stage and encodes the prediction with a two-layer MLP then recalculate the result in the same way in the refinement stage. But this does not produce better cooperation between the two stage. Inspired by SmartRefine\cite{zhou2024smartrefine}, they introduce a brand new framework for refinement and design a quality score mechanism, we design a scoring mechanism between the proposal and refinement stage following HPNet\cite{tang2024hpnet}.\par

\section{Method}
In this section, we first introduce problem formulation for dynamic trajectory prediction in \ref{PF}. In order to verify the performance of the modules we designed and make our network easier to understand, we will introduce the selected backbone network in \ref{HPB}. Then, we present our proposed Graph Attention Mamba Network and the quality scoring mechanism in the two-stage framework in \ref{GAM} and \ref{SM} respectively. Ultimately, we introduce the training objective with the loss function in \ref{TL}.\par

\subsection{Problem Formulation}\label{PF}
The target of trajectory prediction is predicting the future paths of interested agents based on their past movements. Given a fixed-length sequence of history status frames, $\left\{ f_{-T+1}, f_{-T+2},...,f_{0} \right\}$, the goal is to predict K diverse possible trajectories for each of the N agents, as illustrated below:
\begin{equation}
L_0 = \left\{L_{0,n,k}\right\}_{n\in[1,N],k\in[1,K]}\label{eq1}
\end{equation}
where $f_t = \left\{a^{1 \sim N}_t,M\right\}$, $a^{1 \sim N}_t$ represents the features of all agents in the scene at time t, and $M$ denotes the HD map including $N_M$ lane segments. Specifically, $a^n_t=\left\{p^{t,n}_x,p^{t,n}_y,\theta^{t,n},v^{t,n}_x,v^{t,n}_y,c^{t,n}_a\right\}$, where $(p^{t,n}_x,p^{t,n}_y)$ means the location, $\theta^{t,n}$ is the orientation, $(v^{t,n}_x,v^{t,n}_y)$ is the velocity and $c^{t,n}_a$ is the attribute. Every trajectory includes future locations for the next $F$ time steps:
\begin{equation}
L_{0,n,k} = \left\{l_{1,n,k},l_{2,n,k},...l_{F,n,k}\right\}\label{eq2}
\end{equation}
where $l_{i,n,k}\in\mathbb{R}^2$ represents the predicted position at time step $i$ of mode $k$ for agent $n$.\par
\subsection{HPNet Backbone}\label{HPB}
Our work is based on a SOTA approach HPNet\cite{tang2024hpnet}. The encoder is applied by a two-layer MLP, following them, to encode the features of agents and HD map as embeddings:
\begin{equation}
E^{t,n}_a=MLP(v^{t,n},\varphi^{t,n},c^{t,n}_a)\label{eq3}
\end{equation}
\begin{equation}
E_m=MLP(l_m,c_m)\label{eq4}
\end{equation}
where $\varphi^{t,n}$ is the direction of velocity, $l_m$ is the length of lane segments, $c_m$ is the attributes of lane segments, $E^{t,n}_a\in\mathbb{R}^D$, $E_m\in\mathbb{R}^{M \times D}$, and $D$ is the hidden dimension. Each agent at each time step and lane segment are treated as node in the graph. The edge features are represented as $\left\{d_e,\phi_e,\psi_e,\delta_e\right\}$, where $d_e$ denotes the distance between the source and target nodes, $\phi_e$ represents the orientation of the edge in the reference frame of the target node, $\psi_e$ is the relative orientation between source and target nodes, and $\delta_e$ corresponds to the time difference between them. The edge features are encoded into edge embeddings through a two-layer MLP $E_e=MLP(d_e,\phi_e,\psi_e,\delta_e)$, where $E_e\in\mathbb{R}^{Y \times D}$, $Y$ is the number of edges.
The output embeddings from the encoder as the input of Backbone, which contains three main modules driven by our proposed module namely Agent GAM, Historical Prediction GAM and Mode GAM respectively. Agent GAM first input the prediction embeddings $P_{t,n,k}=HP(E_m,E_e,E^{t,n}_a)$, where function $HP$ means the process method in HPNet\cite{tang2024hpnet}, to model the interactions among agents. Then Historical Prediction GAM inputs the result of Agent GAM to model the correlation between historical predictions and current forecast. Finally, results of previous modules are entered into Mode GAM that models interactions among different future trajectory mode and the modules above are repeated $N_{rep} = 2$ times. To further model the sequence relationships, a Mamba block is employed at the end of the three modules.\par
\subsection{Graph Attention Mamba Module with Gate Mechanism}\label{GAM}
An overview of our method is showed in Fig.~\ref{overview}. Our proposed module is applied in the Backbone and it is designed to enhance the feature extraction and prediction capabilities of the network. Specifically, as illustrate in Fig.~\ref{GAMpng}, we use a Mamba2 layer as the Mamba block, graph attention network as GAT block, and the input graph node features and edge features are $P_{t,n,k}$ and $[P_{t,n^{'},k},E_e]$ respectively, where $n^{'}$ represents all agents within a radius of the n-th agent in the same time step and mode. Node features are passed into both Mamba block and GAT block, edge features are just pass through GAT block:
\begin{equation}
P^M_{t,n,k}=P_{t,n,k}+Mam(LN(P_{t,n,k})\label{eq5}
\end{equation}
\begin{equation}
P^A_{t,n,k}=P_{t,n,k}+GAT(LN(P_{t,n,k}),LN(E_e))\label{eq6}
\end{equation}
where function $Mam(a)$ represents the Mamba2 layer, $a$ is the input sequence, $GAT$ function is the graph attention layer, $LN$ means layer normalization and $P^M_{t,n,k}$, $P^A_{t,n,k}$ is the output from the mamba block and GAT block.\par
We also design a gate mechanism to fuse $P^M_{t,n,k}$ and $P^A_{t,n,k}$:
\begin{equation}
G_{t,n,k}=\sigma(F_{fc}(P^M_{t,n,k}+P^A_{t,n,k}))\label{eq7}
\end{equation}
\begin{equation}
P^G_{t,n,k}=P_{t,n,k}+G_{t,n,k} \cdot P^A_{t,n,k} + (1 - G_{t,n,k}) \cdot P^M_{t,n,k}\label{eq8}
\end{equation}
where $F_fc(X)$ is a fully connected layer, $\cdot$ is the sigmoid function, $P^G_{t,n,k}$ is the output of the whole module.\par

\subsection{Quality Scoring Mechanism}\label{SM}
To enhance the performance of two-stage trajectory prediction framework, we introduce a scoring mechanism the evaluates the prediction quality of both the proposal and refinement stages. At the training stage, the quality of predicted trajectory can be assessed according to the ground truth trajectory and the predicted trajectory, inspired by SmartRefine\cite{zhou2024smartrefine}. In detail, using the maximum predicted error between the predicted result and the ground truth, represented by $d_{max}$, calculate the ratio of the absolute value of the difference between the proposal stage and refinement stage result and the absolute value of the difference between the refinement stage result and $d_{max}$ to obtain the quality score:
\begin{equation}
q_{t,n}=\frac{|d_p-d_r|}{|d_{max}-d_r| + \epsilon}\label{eq9}
\end{equation}
where $d_p$ is the predict error at proposal stage, $d_r$ is the predict error at refinement stage. In order to ensure that the calculation is differentiable, we add a very small value $\epsilon$ that is not 0 to the denominator.
To enable GAMDTP to predict the quality score, we utilize a Mamba2 layer to process the prediction embedding at proposal stage. Subsequently, an MLP is employed to produce the quality score, as show in Algorithm ~\ref{score}.
\begin{algorithm}
    \caption{Scoring Mechanism in Training Stage}
    \label{score}
    \renewcommand{\algorithmicrequire}{\textbf{Output:}}
    \renewcommand{\algorithmicensure}{\textbf{Input:}}
    \begin{algorithmic}[1]
        \ENSURE{Proposal section $f_{propose}$, refinement section $f_{refine}$, quality score decoder $f_q$, prediction error function $f_e$, score function $Q$, prediction error function $Dis$ agent embeddings $E_a$, HD map embeddings $E_m$, edge embeddings $E_e$, ground truth trajectory $p_{gt}$, predict agent number $N$}
        \REQUIRE{predicted score $q_p$, calculated score $q$}
        \STATE $d_p, d_r, q, q_p$ $[]$
        \FOR{$t=-T+1,-T+2,...0$}
        \FOR{$n=1,2,...,N$}
        \IF{$n==1 or d_r is None$}
            \STATE $d_{max}=0$
        \ENDIF
        \STATE $p=f_{propose}(E_a,E_m,E_e)$
        \textcolor{blue}{\COMMENT{\% proposal trajectory $p$}}
        \STATE $q_p$ $add$ $f_q(p)$
        \textcolor{blue}{\COMMENT{\% trajectory refinement $\Delta p$}}
        \STATE $\Delta p=f_{refine}(p,E_a,E_m,E_e)$ 
        \STATE $p_{out}=p+\Delta p$
        \STATE $d_p$ $add$ $Dis(p,p_{gt})$ \textcolor{blue}{\COMMENT{\% propose prediction error $d_p$}}
        \STATE $d_r$ $add$ $Dis(p_{out},p_{gt})$ \textcolor{blue}{\COMMENT{\% refine prediction error $d_r$}}
        \STATE $d_{max}=max(d_r, d_{max})$
        \STATE $q$ $add$ $Q(d_{max},d_p,d_r)$
        \ENDFOR
        \ENDFOR
        \RETURN $ q_p, q$
    \end{algorithmic}
\end{algorithm}
\subsection{Training Loss}\label{TL}
To optimize the proposed model, we follow the winner-takes-all strategy, which ensures that the most relevant mode, based on the minimum endpoint displacement, is selected for optimization. Specifically, the $k_{t,n}$-th mode to be optimized is determined by minimizing the endpoint displacement between the predicted trajectory $\left\{L_{t,n,k}\right\},k\in[1,K]$ and the ground truth trajectory $P^{gt}_{t,n}=\left\{p^{gt}_{t+1,n},p^{gt}_{t+2,n},...,p^{gt}_{t+F,n}\right\}$:
\begin{equation}
k_{t,n}=\mathop{\arg\min}_{k\in[1,K]} (l_{t+F,n,k},p^{gt}_{t+F,n})\label{eq10}
\end{equation}
Then two Huber losses are employed to optimize the trajectories both in proposal and refinement stage:
\begin{equation}
\mathcal{L}^{t,n}_{reg1}=\mathcal{L}_{Huber}(L^p_{t,n,k_{t,n}},P^{gt}_{t,n})\label{eq11}
\end{equation}
\begin{equation}
\mathcal{L}^{t,n}_{reg2}=\mathcal{L}_{Huber}(L^r_{t,n,k_{t,n}},P^{gt}_{t,n})\label{eq12}
\end{equation}
where $L^p_{t,n,k_{t,n}}$ is the predicted result in proposal stage, $L^r_{t,n,k_{t,n}}$ is in refinement stage.\par
The probability $\alpha_{t,n,k}$ for each predicted trajectory are optimized using a cross-entropy loss:
\begin{equation}
\mathcal{L}^{t,n}_{cls}=\mathcal{L}_{CE}(\left\{\alpha_{t,n,k}\right\}_{k\in[1,K]},k_{t,n})\label{eq13}
\end{equation}
For the quality scoring mechanism, we calculate the ${\ell_1}$ loss between the predicted score $\hat{q}_{t,n}$ and labeled score $q_{t,n}$:
\begin{equation}
\mathcal{L}_s=\parallel \hat{q}_{t,n}-q_{t,n} \parallel_1\label{eq14}
\end{equation}

\begin{table}[t]
    \caption{Comparison of GAMDTP with the state of the art methods on the Argoverse test set. The b-minFDE is the official ranking metric. For each metric, the best result is in \textbf{bold}, the second best result is \underline{underlined}.}
    \centering
    \label{table1}
    \resizebox{\textwidth}{22mm}{
    \begin{tabular}{l|ccccccc}
    \hline
         $Method$ & $\textbf{b-minFDE}_6\downarrow$ & $minFDE_6\downarrow$ & $minADE_6\downarrow$ & $MR_6\downarrow$ & $minFDE_1\downarrow$ & $minADE_1\downarrow$ & $MR_1\downarrow$ \\ \hline
         LaneGCN\cite{liang2020learning}             & 2.0539 & 1.3622 & 0.8703 & 0.1620 & 3.7624 & 1.7019 & 0.5877 \\  
         mmTransformer\cite{wang2024mm}              & 2.0328 & 1.3383 & 0.8436 & 0.1540 & 4.0033 & 1.7737 & 0.6178 \\
         THOMAS\cite{gilles2021thomas}               & 1.9736 & 1.4388 & 0.9423 & \underline{0.1038} & \underline{3.5930} & 1.6686 & 0.5613 \\
         HOME+GOHOME\cite{gilles2021home}            & 1.8601 & 1.2919 & 0.8904 & \textbf{0.0846} & 3.6810 & 1.6986 & 0.5723 \\
         DenseTNT\cite{gilles2021thomas}             & 1.9759 & 1.2815 & 0.8817 & 0.1258 & 3.6321 & 1.6791 & 0.5843 \\
         MultiModalTransformer\cite{huang2022multi}  & 1.9393 & 1.2905 & 0.8372 & 0.1429 & 3.9007 & 1.7350 & 0.6023 \\
         HiVT\cite{zhou2022hivt}                     & 1.8422 & 1.1693 & 0.7735 & 0.1267 & \textbf{3.5328} & \textbf{1.5984} & \textbf{0.5473} \\
         Mutipath++\cite{varadarajan2022multipath++} & 1.7932 & 1.2144 & 0.7897 & 0.1324 & 3.6141 & \underline{1.6235} & 0.5645 \\
         HPNet(w/o ensemble)\cite{tang2024hpnet}     & \textbf{1.7375} & \textbf{1.0986} & \underline{0.7612} & 0.1067 & 3.7632 & 1.7346 & 0.5514 \\
         \textbf{GAMDTP(ours)}                       & \underline{1.7690} & \underline{1.1256} & \textbf{0.7603} & 0.1088 & 3.8807 & 1.7813 & \underline{0.5509} \\
    \hline
    \end{tabular}}
\end{table}

In summary, final training objective combines the loss functions above:
\begin{equation}
\mathcal{L}=\frac{1}{TN}\sum_{t=-T+1}^{0}\sum_{n=1}^{N}(\mathcal{L}^{t,n}_{reg1}+\mathcal{L}^{t,n}_{reg2}+\mathcal{L}^{t,n}_{cls}+\lambda\cdot\mathcal{L}_s)\label{eq15}
\end{equation}
where $\lambda$ is a hyper-parameter to balance the four loss terms.

\begin{table}[t]
    \centering
    \caption{Comparison of GAMDTP with the state of the art methods on the INTERACTION test set. For each metric, the best result is in \textbf{bold}, the second best result is \underline{underlined}.}
    \centering
    \label{table2}
    \begin{tabular}{l|ccc}
    \hline
         $Method$ & $minJointADE\downarrow$ & $minJointFDE\downarrow$ & $CCR\downarrow$ \\ \hline
         THOMAS\cite{gilles2021thomas}       & 0.4164 & 0.9679 & 0.1791 \\
         DenseTNT\cite{gu2021densetnt}     & 0.4195 & 1.1288 & 0.2240 \\
         Traj-MAE\cite{chen2023traj}     & 0.3066 & 0.9660 & 0.1831 \\
         HDGT\cite{jia2023hdgt}         & 0.3030 & 0.9580 & 0.1938 \\
         FJMP\cite{rowe2023fjmp}         & 0.2752 & 0.9218 & 0.1853 \\
         HPNet(w/o ensemble)\cite{tang2024hpnet}        & \underline{0.2548} & \textbf{0.8231} & \underline{0.1480} \\
         \textbf{GAMDTP(ours)} & \textbf{0.2529} & \underline{0.8295} & \textbf{0.1474} \\
    \hline
    \end{tabular}
\end{table}

\section{Experiments}
\subsection{Datasets}
To evaluate the performance of our model, we conduct experiments on the Argoverse and INTERACTION datasets.\par
\textbf{Argoverse}\cite{chang2019argoverse} is a widely used benchmark for motion forecasting and perception tasks in autonomous driving. It comprises 324,557 interesting vehicle trajectories extracted from over 1,000 driving hours in real-world scenarios. This rich dataset includes high-definition (HD) maps and recordings of sensor data, referred to as “log segments,” collected in two U.S. cities: Miami and Pittsburgh. These cities were chosen for their distinct urban driving challenges, including unique road geometries, local driving habits, and a variety of traffic conditions.\par
\textbf{INTERACTION}\cite{zhan2019interaction} is a comprehensive resource designed to support research in autonomous driving, particularly in behavior-related areas such as motion prediction, behavior cloning, and behavior analysis. It offers a large-scale collection of naturalistic motions from various traffic participants, including vehicles and pedestrians, across a diverse set of highly interactive driving scenarios from different countries.
\subsection{Metrics}
We utilized standard trajectory forecasting metrics, ensuring a comprehensive assessment across different prediction scenarios. These metrics include evaluations on both Argoverse\cite{chang2019argoverse} and INTERACTION\cite{zhan2019interaction} datasets, capturing the accuracy, reliability, and multimodal capabilities of the predictions. For the Argoverse dataset, we employ minimum Average Displacement Error(minADE) and minimum Final Displacement Error(minFDE) to measure the accuracy of trajectory predictions. Specifically, minADE computes the average ${\ell_2}$-norm distance between the predicted trajectory and the ground truth across all time steps, while minFDE focuses on the ${\ell_2}$-norm distance at the final trajectory point. To further assess reliability, we included the Miss Rate(MR), which calculates the proportion of predicted trajectories whose endpoints deviate more than 2.0 meters from the actual ground truth endpoint. Additionally, we employed Brier Minimum Final Displacement Error(b-minFDE), which extends minFDE by integrating a confidence term $(1 - \hat{\alpha})^2$, where $\hat{\alpha}$ represents the predicted probability of the best trajectory. This metric combines endpoint accuracy with the model’s confidence, offering deeper insights into the reliability of its predictions. For the INTERACTION\cite{zhan2019interaction} dataset, we employ minJointADE, minJointFDE and Cross Collision Rate to evaluate the performance of joint trajectory prediction. MinJointADE measures the average $\ell_2$-norm distance between the predicted and ground-truth trajectories of all agents, while minJoint FDE evaluates the $\ell_2$-norm distance at the final time step for all agents. To assess the model's ability to capture multimodal outputs, we set $K = 6$ for both marginal and joint predictions.

\begin{table}[t]
    \centering
    \caption{Ablation study on INTERACTION test set.}
    \centering
    \label{table3}
    \resizebox{\textwidth}{15mm}{
    \begin{tabular}{cccccc|ccccc}
    \hline
         $Backbone$ & $Gate$ & $Score$ & $1-layer$ & $3-layers$ & $5-layers$ & $minJointADE\downarrow$ & $minJointFDE\downarrow$ & $CCR\downarrow$ & $minJointMR\downarrow$ \\ \hline
         \checkmark &     &       & \checkmark &       &          & 0.2641 & 0.8610 & 0.1515 & 0.1717 \\
         \checkmark & \checkmark & & \checkmark & & & \underline{0.2543} & \underline{0.8342} & \underline{0.1473} & \underline{0.1530} \\
         \checkmark & & \checkmark & \checkmark & & & 0.2641 & 0.8614 & 0.1516 & 0.1700 \\
         \checkmark & \checkmark & \checkmark & \checkmark & & & \textbf{0.2529} & \textbf{0.8295} & \textbf{0.1474} & \textbf{0.1525} \\
         \checkmark & \checkmark & \checkmark & & \checkmark & & 0.2643 & 0.8614 & 0.1511 & 0.1722 \\
         \checkmark & \checkmark & \checkmark & & & \checkmark & 0.2706 & 0.8687 & 0.1548 & 0.1665 \\
    \hline
    \end{tabular}}
\end{table}
\subsection{Comparison with State-of-the-art}
\begin{figure*}[t]
	\centering
        \subfigure baseline(w/o ensembling){
        \begin{minipage}[t]{1\linewidth}
        \centering
        \includegraphics[width=1\linewidth]{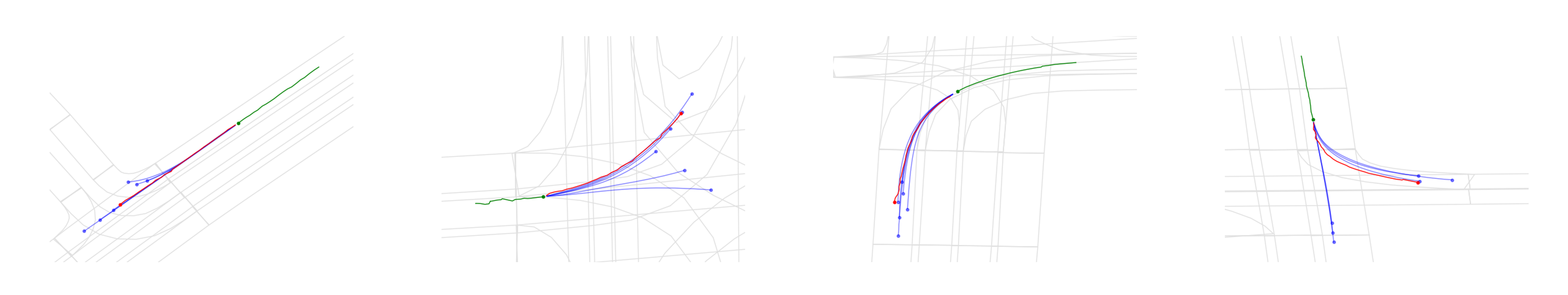}
        \end{minipage}
        }
        \subfigure GAMDTP(ours){
        \begin{minipage}[t]{1\linewidth}
            \centering
            \includegraphics[width=1\linewidth]{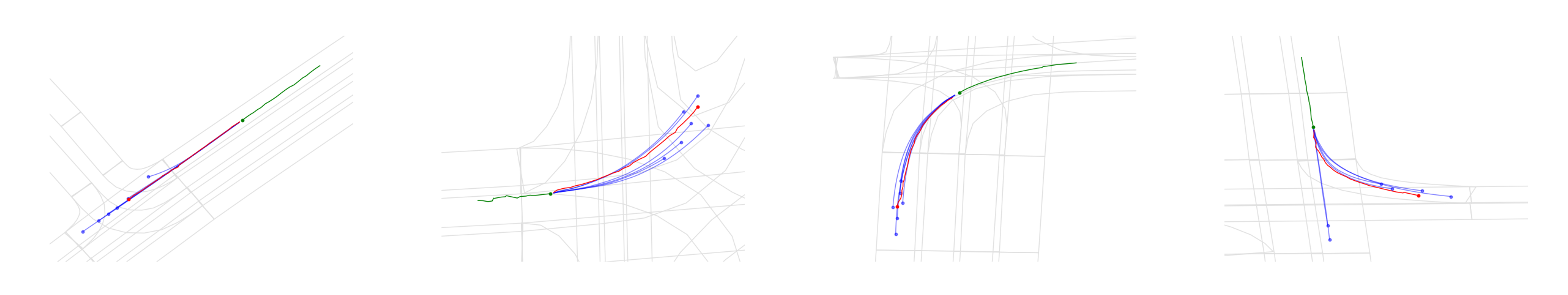}
        \end{minipage}
        }
        \caption{Comparison our GAMDTP with baseline.}
        \label{comparison}
\end{figure*}

\textbf{Results on Argoverse.} The results for marginal trajectory prediction on the Argoverse\cite{chang2019argoverse} dataset are presented in Table ~\ref{table1}. Our GAMDTP achieves the SOTA performance across all evaluation metrics among single models. Compared to HPNet\cite{tang2024hpnet}, the second-best model on Argoverse leaderboard, GAMDTP improves from 0.7612 to 0.7603 in minADE where mode number $K = 6$ and from 0.5514 to 0.5509 in MR where mode number $K = 1$. These improvements highlight the effectiveness of our model in accurately capturing both trajectory endpoints and multimodal predictions. We show some examples in Fig.~\ref{comparison}.\par
\textbf{Results on INTERACTION.} Table ~\ref{table2} presents the performance of our method on the INTERACTION\cite{zhan2019interaction} multi-agent track, where we achieved state-of-the-art results. Our approach outperformed the first-ranked FJMP\cite{rowe2023fjmp} by a significant margin, with improvements of 0.0223 in minJointADE, 0.0923 in minJointFDE and 0.0379 in Cross Collision Rate(CCR). Improve about 4\% in CCR and 0.0019 in minJointADE compared to backbone HPNet\cite{tang2024hpnet}. These results demonstrate that our GAMDTP is both a simple and effective solution for joint trajectory prediction. 

\subsection{Ablation Study}
To check the effectiveness of the key components in our model, we conduct a series of ablation experiments on the INTERACTION\cite{zhan2019interaction} test set. Specifically, we evaluate the impact of the gate mechanism, quality scoring mechanism and the number of Mamba layers. The results are summarized in Table ~\ref{table3}.\par

\textbf{Effect of gate mechanism.} The gate mechanism in our GAMDTP module dynamically balances the contributions of the GAT and Mamba-SSM outputs. To analyze its effectiveness, we compare the model’s performance with and without the gate mechanism. As shown in Table ~\ref{table3}, the removal of the gate mechanism leads to a noticeable drop in performance, increase minJointADE from 0.2529 to 0.2641, minJointFDE from 0.8295 to 0.8614 and the Cross Collision Rate(CCR) from 0.1474 to 0.1516. These results highlight the importance of dynamically balancing the contributions of GAT and Mamba-SSM for effective feature extraction. \par
\textbf{Effect of score mechanism.} The quality scoring mechanism evaluates the reliability of trajectory proposals and guides the refinement process. To evaluate its impact, we compare the model with and without this mechanism. The absence of the scoring mechanism results in an increase of 0.0014 in minJointADE, 0.0047 in minJointFDE and CCR slightly increases from 0.1473 to 0.1474. This demonstrates that the quality scoring mechanism effectively enhances the refinement process by prioritizing reliable trajectories.\par
\textbf{Effect of different numbers of Mamba layers.} We investigate the impact of varying the number of Mamba layers in GAMDTP. Specifically, we test configurations with 1, 3 and 5 layers. The results in Table ~\ref{table3} indicates that more Mamba layers will lead to performance degradation, the minJointADE increases from 0.2529 to 0.2643 in 3 layers and 0.2706 in 5 layers respectively, the minJointFDE increases from 0.8295 to 0.8614 and 0.8687 and the CCR increases from 0.1474 to 0.1511 and 0.1548, this likely that more layers lead to computational redundancy, resulting in difficulty in convergence. To balance comprehensive performance and computational efficiency, we choose 1 Mamba layer in our GAMDTP network.
\section{Conclusion}
In this paper, we introduced GAMDTP, a novel framework for accurate and efficient trajectory forecasting in autonomous driving scenarios. By integrating Mamba-SSM and Graph Attention Networks (GAT) through a dynamic gating mechanism, our model effectively captures both local spatial interactions and global temporal dependencies. To further enhance the two-stage trajectory prediction framework, we designed a Quality Scoring Mechanism, which evaluates trajectory proposals and prioritizes high-quality candidates during refinement. Our experimental results on the Argoverse and INTERACTION datasets demonstrate that GAMDTP achieves state-of-the-art performance. In summary, GAMDTP offers a scalable and reliable solution for dynamic trajectory forecasting, advancing the capabilities of autonomous driving systems.

%
%
%




\bibliographystyle{ieeetr}
\bibliography{IEEEabrv,mylib}
\end{document}